\pdfoutput=1

\documentclass[11pt]{article}

\usepackage[]{acl}

\usepackage{times}
\usepackage{latexsym}
\usepackage{todonotes}

\usepackage[T1]{fontenc}

\usepackage[utf8]{inputenc}

\usepackage{microtype}

\usepackage{inconsolata}

\usepackage{tabularx}
\usepackage{amsmath}
\usepackage{amsfonts}
\usepackage{amssymb}
\usepackage{color}
\usepackage{tabu}
\usepackage{multirow}
\usepackage{multicol}
\usepackage{colortbl}
\usepackage{subcaption}
\usepackage{graphicx}
\usepackage{caption}
\usepackage{hhline}
\usepackage{booktabs}
\usepackage{enumitem}
\usepackage{tabularx}

\usepackage{xspace}
\newcommand\datasetname{\mbox{\textbf{\textsc{ACLSum}}}\xspace}

\title{ACLSum: A New Dataset for\\%
Aspect-based Summarization of Scientific Publications}

\author{Sotaro Takeshita\textsuperscript{1}, Tommaso Green\textsuperscript{1}, Ines Reinig\textsuperscript{1},\\\textbf{Kai Eckert\textsuperscript{2}, Simone Paolo Ponzetto\textsuperscript{1}}\\
  \textsuperscript{1}Data and Web Science Group, University of Mannheim, Germany \\
  \textsuperscript{2}Mannheim University of Applied Sciences, Mannheim, Germany \\ 
  \texttt{\{sotaro.takeshita, tommaso.green, ines.reinig, ponzetto\}@uni-mannheim.de} \\
  \texttt{k.eckert@hs-mannheim.de}}

\begin{document}
\maketitle

\begin{abstract}
Extensive efforts in the past have been directed toward the development of summarization datasets. However, a predominant number of these resources have been (semi)-automatically generated, typically through web data crawling, resulting in subpar resources for training and evaluating summarization systems, a quality compromise that is arguably due to the substantial costs associated with generating ground-truth summaries, particularly for diverse languages and specialized domains. To address this issue, we present \datasetname, a novel summarization dataset carefully crafted and evaluated by domain experts. In contrast to previous datasets, \datasetname  facilitates multi-aspect summarization of scientific papers, covering challenges, approaches, and outcomes in depth. Through extensive experiments, we evaluate the quality of our resource and the performance of models based on pretrained language models and state-of-the-art large language models (LLMs). Additionally, we explore the effectiveness of extractive versus abstractive summarization within the scholarly domain on the basis of automatically discovered aspects. Our results corroborate previous findings in the general domain and indicate the general superiority of end-to-end aspect-based summarization.\footnote{Our data is released at \url{https://github.com/sobamchan/aclsum}.}
\end{abstract}

\section{Introduction}
The availability of high-quality datasets annotated with ground-truth human judgements has been a staple of the elements required to advance research in NLP for a very long time, dating back to the very dawn of statistical NLP \cite{marcus-etal-1993-building}. Unfortunately, the availability of such resources has been quite scarce in the domain of text summarization of scientific papers \cite{koh-etal-2022-empirical}. A prevalent approach in summarization from the last few years is to semi-automatically collect text snippets from the Internet that would serve as pseudo-summaries, with only partial quality control \citep{kryscinski-etal-2019-neural,tejaswin-etal-2021-well}. While this enabled the creation of large datasets for data-hungry learning methods, it also makes it challenging to truly capture the summarization capabilities of models, especially when viewed together with the limitations of existing benchmarking evaluation metrics, e.g., ROUGE \citep{lin_rouge_2004} and the widespread use of incorrect implementations of such metrics within the community \cite{grusky-2023-rogue}.
\begin{figure}
    \centering
    \small
    \includegraphics[width=\columnwidth]{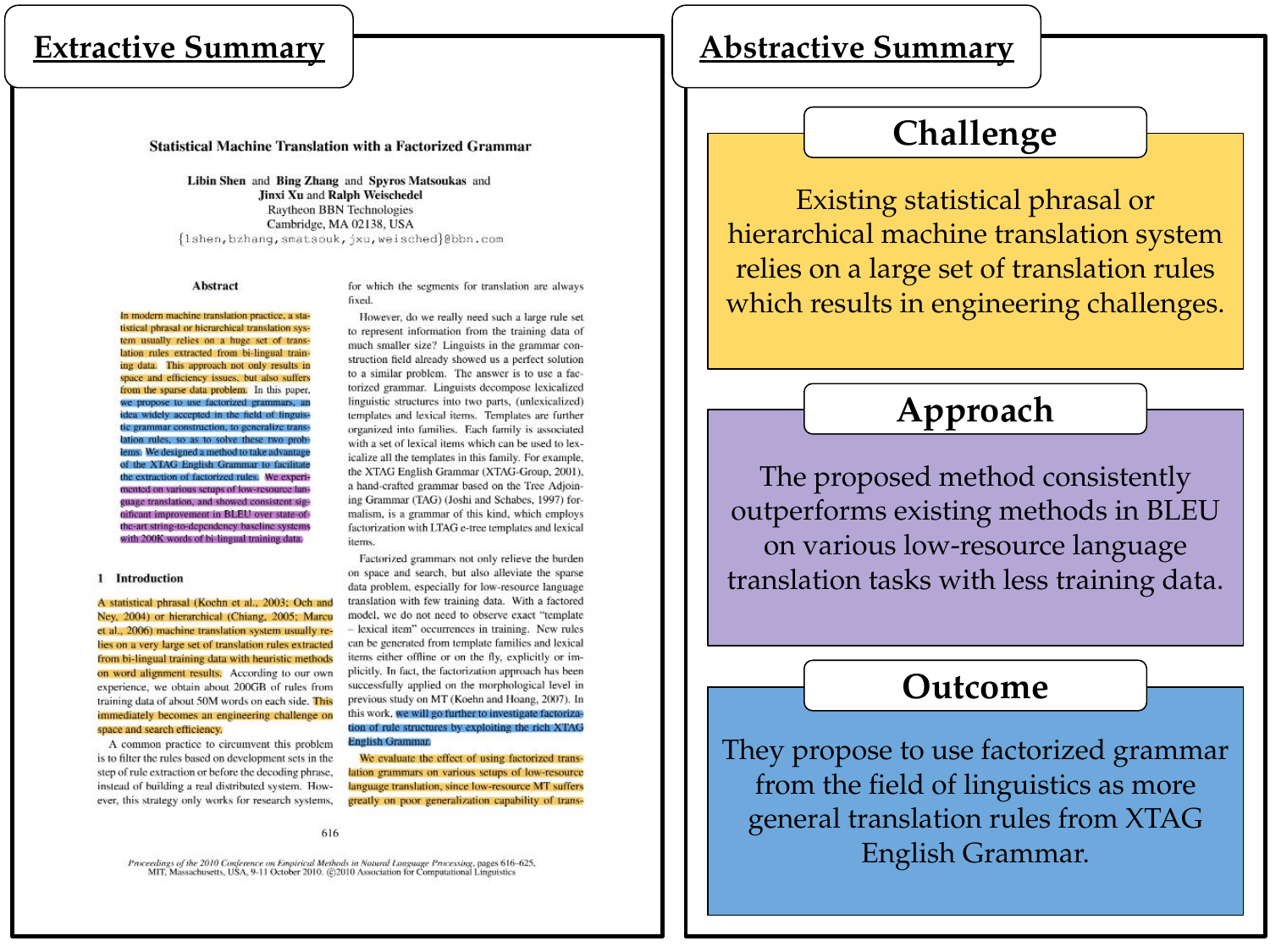}
    \caption{A data sample from \datasetname. Each document is complemented with manually-crafted and validated summaries for both extractive and abstractive setups on three different aspects. We annotate aspects to be used as extractive summaries.}
    \label{fig:main-fig}
\end{figure}
This vicious interplay of partially reliable evaluation metrics and semi-automatically extracted data has also been observed in machine translation: \citet{freitag_bleu_2020} shows that the low correlation between human and automatic evaluation is not only caused by the nature of the evaluation metrics but also by the lack of proper reference translations. In summarization, \citet{zhang_benchmarking_2023} showed that the agreement between an evaluation metric and human judgements can be increased by improving reference summaries with no modification to the metric itself. While new metrics have been proposed to improve the correlation to human preference and replace ROUGE \citep{zhang_bertscore_2019,deutsch_towards_2021}, only a few works tackle the reliability of evaluating summarization from the standpoint of dataset quality. This is especially true for the domain on which we focus in this paper, namely the automatic multi-aspect summarization of scientific papers, where we want to preserve output truthfulness to produce a correct summary along three different aspects of scientific work: the main challenge addressed in the paper, the approach developed to overcome it and the overall outcome of the work. 

To address these limitations, we present \datasetname, a new multi-aspect extreme summarization dataset manually annotated and validated by domain experts. We propose a two-stage summary annotation approach where, for each of the proposed aspects, the annotators first select aspect-relevant sentences in the source documents and then use these to produce an abstractive summary. The process is represented in Figure~\ref{fig:main-fig}: the left-hand side contains the paper with the color-coded relevant sentences for each of the aspects, while the right-hand side contains summaries for each aspect. 
This results in each source document having two kinds of gold standard annotations, namely (1) the set of sentences relevant to each of the aspects and (2) abstractive reference summaries. We evaluate the quality of our dataset through manual validation from domain experts.

Using \datasetname, we perform three lines of experiments to evaluate different summarization strategies. First, we compare two approaches for text summarization with pretrained language models (PLMs): (i) end-to-end summarization, in which the PLM directly produces a summary from the source document, and (ii) extract-then-abstract summarization, in which an extractive model first extracts sentences that are then used as input to the PLM to generate the summary. The unique property of our dataset is having gold annotations both for aspects and summaries. This enables a fine-grained analysis: we quantitatively show that generative models suffer more when the relevant information is scattered across the source document, thus requiring them to perform a higher level of abstraction to produce final summaries.
Second, we shed light on recently developed large language models (LLMs) by training and evaluating Llama 2 \citep{touvron_llama_2023} in two different ways, namely through (i) end-to-end instruction-tuning where we train the model to produce summaries directly given an instruction sentence and a source document as an input; and (ii) extract-then-abstract chain-of-thought-like training, where we instruct-tune the model first to generate references to the sentences in the source document that cover relevant aspects for the summary, and then merge these sentences to produce a final summary. Third, we evaluate a greedy algorithm used in previous work to induce silver-standard extractive summaries \citep{nallapati2017summarunner} and empirically show its low quality when properly evaluated against ground-truth annotations from human experts.
Our contributions are the following ones:
\begin{enumerate}[itemsep=0mm,leftmargin=4mm]
    \item \textbf{A new expert-annotated and validated multi-aspect summarization dataset} with both extractive and extreme abstractive summary annotations.
    \item \textbf{An extensive and fine-grained evaluation} of PLM systems and instruction-tuned LLMs on aspect-based summarization of scientific papers. 
    \item \textbf{A benchmarking assessment of a greedy search heuristic} for extractive summary generation on our domain.
\end{enumerate}
\section{Related Work}
\label{se:problems}
A common practice to build summarization datasets is to find data on the Internet which can be used as a silver-standard proxy for document-summary pairs \citep{cohan-etal-2018-discourse,kim_abstractive_2019,hayashi_wikiasp_2021}, e.g., news articles and their lead sentences \citep{hermann_teaching_2015,narayan_dont_2018}. While scalable and very practical to build datasets on a large scale, the resulting summaries exhibit a few fundamental limitations.

\paragraph{Unfaithful summaries.}
\citet{maynez_faithfulness_2020} show that XSum contains summaries that are unfaithful to source documents in the sense that these `summaries' are rather `eye-catching' sentences to draw readers to the corresponding articles for the news platform from which the dataset is extracted.

\paragraph{Noisy data.}
\citet{kryscinski-etal-2019-neural} report that CNN/DailyMail \citep{hermann_teaching_2015} and Newsroom \citep{grusky-etal-2018-newsroom} contain much noise, such as URLs and placeholder texts in their summaries. More recently, \citet{koh-etal-2022-empirical} show that more than 60\% of the reference summaries in the test set of the arXiv dataset which is built for long document summarization contain noises. Moreover, in a  preliminary study, we investigated automatic language detection models
and found that 0.4\% samples in the test set of XSum \citep{narayan_dont_2018} are, for instance, written in Welsh (as opposed to English).

\paragraph{Legal issues.}
The two most well-used datasets, namely CNN/DailyMail and XSum, raise various legal issues (e.g., copyrighted content) and are not publicly available \citep{wang_squality_2022}.

\paragraph{Missing gold extractive labels.}
While extractive and aspect-based summarization are both active research subjects, there are no freely available datasets with ground-truth labels for such tasks. For extractive summarization, the \textit{de facto} standard approach has been relying on a heuristic-based algorithm that automatically induces labels from abstractive summarization datasets without validating its effectiveness \citep{nallapati2017summarunner}, including recent improvements from \citet{xu2022text}.

\vspace{1em}
\noindent
The work closest to ours is \mbox{SQuALITY} from \citet{wang_squality_2022}. While this work shares the core motivation with our work, which is to build a reliable and validated summarization dataset, our dataset has several different properties. First, besides the abstractive reference summaries, our dataset also has passage annotations (i.e., aspects) that can serve as gold labels for extractive summarization. Second, in contrast to the \mbox{SQuALITY}, which provides question-focused summaries, our dataset has multi-aspect summaries more suitable for our target scholarly domain. Third, \mbox{SQuALITY} uses novel stories as its source documents, whereas our dataset uses research articles from the field of NLP, which makes our dataset highly domain-specific and challenging. Lastly, the size, our \datasetname contains 250 documents, which is more than twice larger than 100 documents from \mbox{SQuALITY}. Another work similar to ours is SciTLDR \citep{cachola-etal-2020-tldr}, a collection of papers from computer science and one-sentence summaries, later extended by \citet{takeshita22} for cross-lingual summarization. This work has inspired us to design reference summaries in our dataset to have one-sentence summaries. Our work differs from theirs in (i) the structure of summaries: ours has multi-aspect summaries instead of one overview summary, (ii) types of annotations: while the SciTLDR contains only abstractive reference summaries, our dataset also contains annotations of the relevant sentences.
\section{Dataset creation}

\paragraph{Source documents.}
We focus on the summarization of scholarly documents \cite{erera_summarization_2019,fok_scim_2022} with a focus on NLP papers because of the availability of large amounts of freely available text \cite{bird-etal-2008-acl} in a challenging domain setting. The focus on NLP is additionally driven by the surge in recent years of publications in our field, and the consequent need for summarization systems, as well as the availability of domain-expert annotators at our disposal.

We take research papers published in five major NLP conferences, namely ACL, NAACL, EMNLP, ECAL, and AACL from 1974 to 2022, and use Grobid \citep{grobid} to extract Abstract, Introduction, and Conclusion (AIC) sections from PDF files.  We apply a bucket-based sampling for selecting documents to be annotated to maintain the diversity of documents in our dataset. Due to the increasing number of published papers in the last decade, random sampling would be biased towards recent publications. To avoid this, we divide the papers into different buckets for each combination of year and venue and uniformly sample from them to create a pool of papers to be annotated.

\paragraph{Summary aspects.}
Recently, there have been several works proposing datasets with multiple summaries for each document to cover different aspects of source documents \citep{hayashi_wikiasp_2021,yang-etal-2023-oasum}. Research articles are also multi-faceted documents with multiple aspects \citep{fisas_discoursive_2015}.
Consequently, our annotated passages and abstractive summaries cover three different aspects, namely: (i) Challenge: \textit{The current situation faced by the researcher; it will normally include a Problem Statement, the Motivation, a Hypothesis and/or a Goal.}, (ii) Approach: \textit{How they intend to carry out the investigation, comments on a theoretical model or framework.}, and finally (iii) Outcome: \textit{Overall conclusion that should reject or support the research hypothesis.}. We operationalize the definition of these three aspects using the definitions from \citet{fisas_discoursive_2015}.

\paragraph{Annotation process.}
We annotate the dataset by relying on domain experts (graduate students in NLP), as opposed to relying on crowd-sourcing platforms, which are known to have quality issues \citep{zhang_needle_2023}. We use a two-stage process to produce reference summaries. In the first stage, we review each sentence in the source document and annotate it with an aspect if it contains information that is relevant to (one of the aspects of) the summary. In the second stage, the annotator writes a summary using selected sentences with a 25-word limit, to maintain the average sentence length of the source document. This property makes \datasetname an extreme summarization dataset, a more challenging setup \citep{narayan_dont_2018} than traditional summarization, which is suitable for the scholarly domain since researchers need to consume a steadily increasing number of papers \citep{bornmann_growth_2015}. We do not make use of any models or systems for our annotation task to avoid any biases that could favor certain models in evaluation \citep{deutsch-etal-2022-limitations}. The full annotation guidelines can be found in the Appendix \ref{sec:summary_guideline}.
\section{\datasetname}
\begin{table*}[t]
    \centering
    \small
    \setlength{\tabcolsep}{2pt}
    \begin{tabular}{rrrrrrrrrrrrrrrrrr}
    \toprule
     & & \multicolumn{4}{c}{\textbf{Document}} & & \multicolumn{4}{c}{\textbf{Aspects}} & & \multicolumn{5}{c}{\textbf{Summaries}}  \\
    \cmidrule{3-6} \cmidrule{8-11} \cmidrule{13-17}
    \multicolumn{1}{l}{\textbf{Aspect}} & & \multicolumn{1}{c}{\begin{tabular}[c]{@{}c@{}}\textbf{\# doc.}\\\textbf{train/val/test}\end{tabular}} & \begin{tabular}[c]{@{}c@{}}\textbf{avg. \#}\\\textbf{words}\end{tabular} & \begin{tabular}[c]{@{}c@{}}\textbf{avg. \#}\\\textbf{sent.}\end{tabular} & \begin{tabular}[c]{@{}c@{}}\textbf{\#}\\\textbf{vocab}\end{tabular} & & \begin{tabular}[c]{@{}c@{}}\textbf{avg. \#}\\\textbf{sent.} \end{tabular} & \begin{tabular}[c]{@{}c@{}}\textbf{avg. \#}\\\textbf{words}\end{tabular} & \begin{tabular}[c]{@{}c@{}}\textbf{\#}\\\textbf{vocab}\end{tabular} & \begin{tabular}[c]{@{}c@{}}\textbf{comp.}\\\textbf{ratio}\end{tabular} & & \begin{tabular}[c]{@{}c@{}}\textbf{avg. \#}\\\textbf{words}\end{tabular} & \begin{tabular}[c]{@{}c@{}}\textbf{avg. \#}\\\textbf{vocab}\end{tabular} & \begin{tabular}[c]{@{}c@{}}\textbf{avg. \#}\\\textbf{new}\\\textbf{words}\end{tabular} & \begin{tabular}[c]{@{}c@{}}\textbf{comp.}\\\textbf{ratio}\\\textbf{(to doc.)}\end{tabular} & \begin{tabular}[c]{@{}c@{}}\textbf{comp.}\\\textbf{ratio}\\\textbf{(to asp.)} \end{tabular} \\
    \midrule
    \multicolumn{1}{l}{\textbf{Challenge}} & & \multirow{3}{*}{100/50/100} & \multirow{3}{*}{914.7} & \multirow{3}{*}{38.45} & \multirow{3}{*}{14k} & & 4.3 & 109.0 & 4.5k & 8.4 & & 22.5 & 1.8k & 3.3 & 40.1 & 4.8 \\
    \multicolumn{1}{l}{\textbf{Approach}} & & & & & & & 6.6 & 162.6 & 4.8k & 5.6 & & 22.7 & 1.7k & 2.1 & 40.1 & 7.1 \\
    \multicolumn{1}{l}{\textbf{Outcome}} & & & & & & & 4.4 & 110.3 & 3.9k & 8.3 & & 21.3 & 1.4k & 2.2 & 42.6 & 5.1 \\
    \bottomrule
    \end{tabular}
    \caption{Statistics of \datasetname\ including documents, sentences annotated with aspects and summaries.}
    \label{ta:dataset}
\end{table*}

Table \ref{ta:dataset} presents statistics of our dataset. \datasetname\ consists of 250 documents (i.e., AICs) with an average length of approximately 40 sentences and 1,000 words. The average length of the annotated aspects is comparable, with passages describing approaches being slightly longer. The compression ratios based on relevant aspect sentences (namely, the average number of words of AICs to the average number of words per annotated aspect, column 9) ranges between 8.4 and 5.6, whereas for abstractive summaries it ranges between 40.1 and 42.6, thus exhibiting the high level of abstraction required for models to perform abstractive summarization.

\begin{table}[t]
    \centering
    \small
    \setlength{\tabcolsep}{8pt}
    \begin{tabular}{rrrr}
    \toprule
    \multicolumn{1}{c|}{\textbf{Aspect}} & \multicolumn{1}{c}{\textbf{Relevance}} & \multicolumn{1}{c}{\textbf{Consistency}} & \multicolumn{1}{c}{\textbf{Fluency}} \\
    \midrule
    \multicolumn{1}{c|}{\textbf{Challenge}} & 4.98 & 4.85 & 4.65  \\
    \multicolumn{1}{c|}{\textbf{Approach}} & 4.82 & 4.70 & 4.56 \\
    \multicolumn{1}{c|}{\textbf{Outcome}} & 4.96 & 4.74 & 4.54 \\
    \bottomrule
    \end{tabular}
    \caption{Manual validation of different aspects along three criteria on a 1-5 Likert scale.}
    \label{tab:validation}
\end{table}

We validate the quality of the annotations in our dataset by taking 75 summaries (25 AICs times three aspects) from our validation split and let two additional domain experts evaluate the quality according to three criteria proposed by \citet{fabbri_summeval_2021}, namely relevance, consistency, and fluency. We do not evaluate \textit{coherence} since this measures ``the quality of all sentences collectively'' \citep{fabbri_summeval_2021} while our reference summaries are composed of a single sentence. The complete annotation guidelines can be found in the Appendix \ref{sec:validation_guideline}. 
On all aspects and criteria, our reference summaries achieve above 4 on a 1-5 Likert scale (Table \ref{tab:validation}), and especially high scores in  Relevance and Consistency show that our reference summaries capture essential information in the source documents. 
We measure inter-annotator agreement between both annotators in terms of percentage agreement. 
The agreement on Relevance is remarkably high (96\% for Challenge and Outcome, 76\% for Approach) and satisfactory for Consistency (68\%, 72\% and 76\% for Challenge, Approach and Outcome respectively). On Fluency however, annotators agreed less frequently for all three aspects (52\%, 52\% and 48\%), which may suggest a more subjective nature of the Fluency measure. 

\begin{figure*}[ht]
    \centering
    \begin{subfigure}[b]{0.5\columnwidth}
        \centering
        \includegraphics[width=\columnwidth]{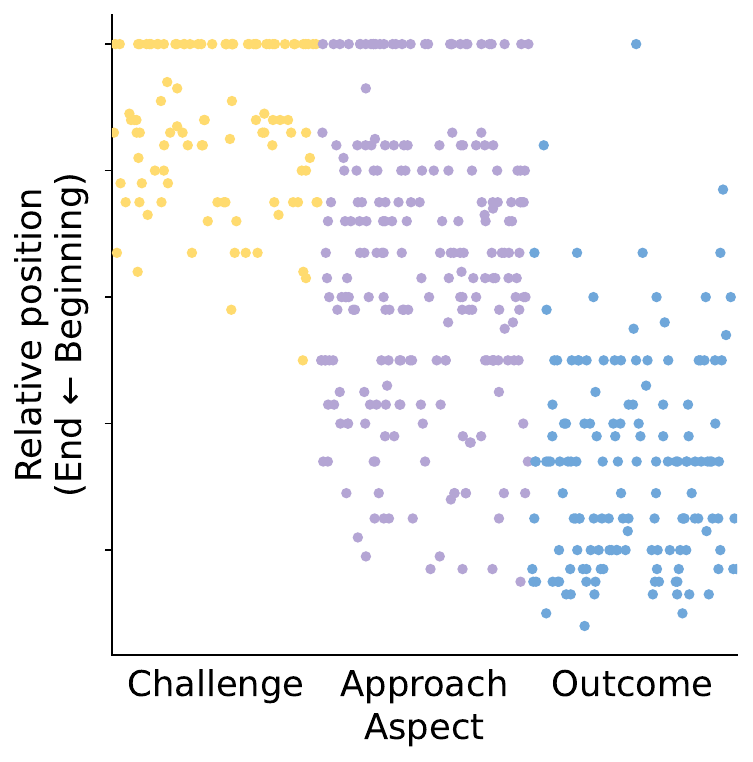}
        \caption{Abstract}
    \end{subfigure}
    \hfill
    \begin{subfigure}[b]{0.5\columnwidth}
        \centering
        \includegraphics[width=\columnwidth]{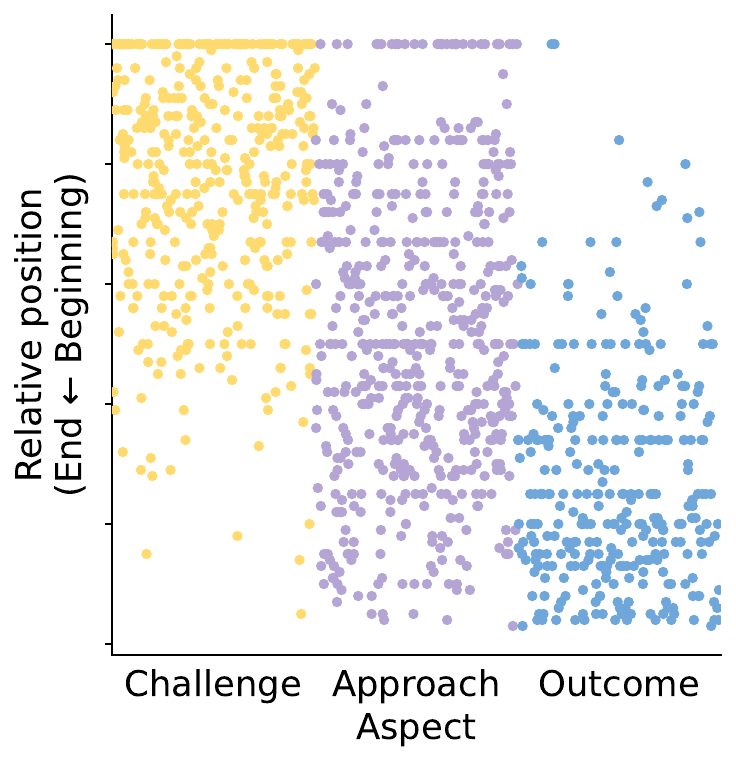}
        \caption{Introduction}
    \end{subfigure}
    \hfill
    \begin{subfigure}[b]{0.5\columnwidth}
        \centering
        \includegraphics[width=\columnwidth]{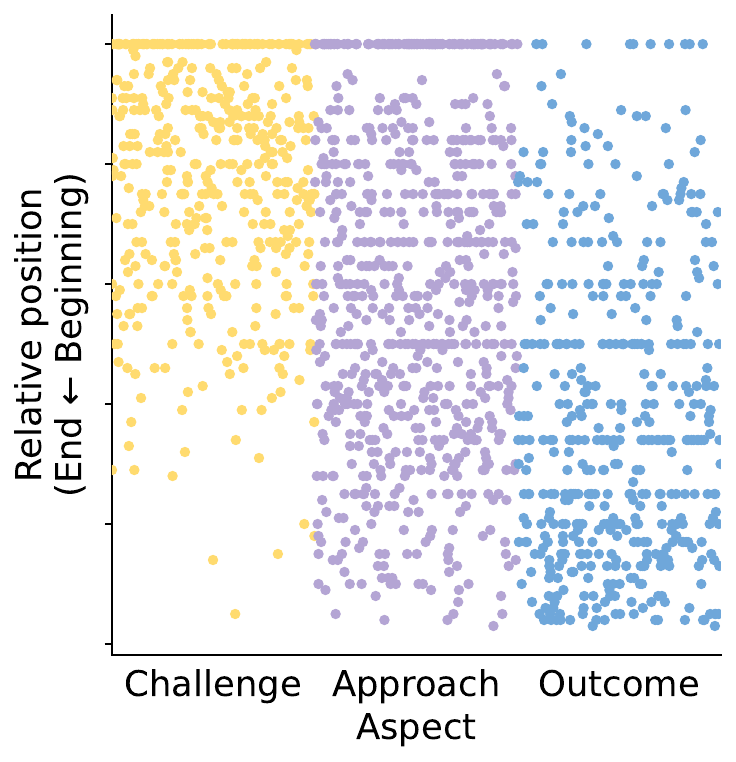}
        \caption{Conclusion}
    \end{subfigure}
    \caption{Relative positions of relevant sentences for each aspect (Challenge, Approach and Outcome).}
    \label{fig:relative-positions}
\end{figure*}

We show in Figure \ref{fig:relative-positions} the relative positions of sentences annotated with aspects, highlighting how the three different aspects are highly interspersed across documents, thus indicating that indeed models are required to attend to many different parts of the document for each aspect.
\section{Experiments and Results}
\label{se:experiments}
We next use \datasetname to answer the following research questions:

\begin{itemize}[itemsep=0mm,leftmargin=4mm]
    \item \textbf{RQ1}: Which approach using PLMs, i.e., two-stage extract-then-abstract or end-to-end abstractive summarization, performs best on our dataset?
    \item \textbf{RQ2}: Which tuning strategy for LLMs, i.e., two-step chain-of-thought or end-to-end abstractive summarization, is better for our task?
    \item \textbf{RQ3}: How does a commonly used heuristic to induce silver-standard extractive summaries perform against our manually annotated aspects?
\end{itemize}

\subsection{RQ1: Extract-then-abstract vs. end-to-end}
\label{sec:rq1}
The extract-then-abstract approach (EtA) has recently become more widely used in the literature \citep{hsu-etal-2018-unified,mao-etal-2022-dyle}. It consists of first using an extractive model to select relevant sentences in the source documents and then deploying an abstractive model to merge extracted sentences into a summary: this is opposed to end-to-end summarization (E2E) in which the model directly generates the summary using the entire source document as input \citep{liu-etal-2022-brio,zhang-etal-2020-pegasus}. While the extract-then-abstract approach potentially suffers from the error propagation problem, it can benefit from more efficient inference (due to the reduced number of sentences fed to the abstractive models) and, arguably, transparency on the provenance of the summary (since summaries are typically generated from a few extracted sentences). In our setting, \datasetname enables ground-truth evaluation of both stages since our dataset contains both annotated aspects (which can be used to provide extractive summaries) and abstractive summaries.

\paragraph{Experimental setup.}
For extractive models, we evaluate the Sentence-T5 model proposed by \citet{ni_sentence-t5_2021} in three different sizes (\textsc{BASE}, \textsc{LARGE}, \textsc{XL}) since it was shown to be a powerful text encoder in a recent study by \citet{muennighoff_mteb_2023}. We use the ST5-Enc mean variant, which only uses the encoder of T5 and applies mean-pooling to get sentence representations. We train a binary logistic regression model on top of Sentence-T5 representations to classify if the text is relevant or irrelevant to producing the summary for the aspect at hand. The annotations for each sentence act as labels (positive class if the sentence was selected by the annotators, negative otherwise.). For the abstractive model, we evaluate BART \citep{lewis_bart_2020} and T5 \citep{raffel_exploring_2020} in two sizes (\textsc{BASE}, \textsc{LARGE}). Additionally, we evaluate the setup -- which we refer to hereafter as `Gold' -- for which we feed manually annotated gold extractive summaries to the abstractive model. This lets us evaluate the upper-bound capability of the abstractive model within the pipeline in isolation. In the E2E setup, i.e., direct summarization, the abstractive model takes the entire document as input. We carry out separate training procedures for each aspect for both approaches.

We train the binary logistic regression classifier with L2 regularization with $C=1.0$ regularization strength.
For each possible choice of extractive model in the EtA setup, we use its predictions to both train and evaluate the abstractive model that follows in the pipeline. For all abstractive models, we perform grid-search using the following grid: learning rate $lr \in$ \{1e-5, 3e-5, 5e-5\}, batch size $B \in \{4, 8, 16\}$ for $\text{BART}_{\textsc{BASE}/\textsc{LARGE}}$ and $B \in \{2, 4, 8\} $ for $\text{T5}_{\textsc{BASE}/\textsc{LARGE}}$. During the hyperparameter search, we use the validation split of our dataset with a fixed seed to find the best combination of parameters according to the loss. We report each score as an average over three differently seeded models. For evaluation, we report standard ROUGE scores \citep{lin_rouge_2004} and BERTScore \citep{zhang_bertscore_2019} with SciBERT \citep{beltagy-etal-2019-scibert} as the underlying model.

\paragraph{Results and discussions.}
\begin{table}[t]
    \centering
    \small
    \setlength{\tabcolsep}{1.8pt}
    \begin{tabular}{lccccccccccc}
    \toprule
     & \multicolumn{3}{c}{\textbf{Challenge}} & & \multicolumn{3}{c}{\textbf{Approach}} & & \multicolumn{3}{c}{\textbf{Outcome}} \\
    \cmidrule{2-4} \cmidrule{6-8} \cmidrule{10-12}
    \textbf{Model} & \textbf{P} & \textbf{R} & \textbf{F1} & & \textbf{P} & \textbf{R} & \textbf{F1} & & \textbf{P} & \textbf{R} & \textbf{F1} \\
    \midrule
    \textbf{$\text{ST5}_{\textsc{BASE}}$}  & 63.2 & 75.7 & 65.0 & & 68.0 & 77.6 & 70.0 & & 71.2 & 84.1 & 74.9 \\
    \textbf{$\text{ST5}_{\textsc{LARGE}}$} & 63.6 & 75.5 & 65.7 & & 69.7 & 79.7 & 72.1 & & 71.5 & 83.9 & 75.2 \\
    \textbf{$\text{ST5}_{\textsc{XL}}$}    & \textbf{63.9} & \textbf{76.3} & \textbf{66.0} & & \textbf{70.3} & \textbf{80.0} & \textbf{72.7} & & \textbf{71.8} & \textbf{84.8} & \textbf{75.6} \\
    \bottomrule
    \end{tabular}
    \caption{Performance of Sentence-T5 in three different sizes on each aspect of our dataset with three metrics, \textbf{F1}, (\textbf{P})recision, and (\textbf{R})ecall.}
    \label{ta:classification}
\end{table}

Table \ref{ta:classification} shows the performance of the extractive models. As the size of the underlying model increases, the performance on all three aspects improves. All models perform worse on the Challenge aspect compared to the other two, possibly because of the fewer annotated passages for this aspect (cf.\ Table~\ref{ta:dataset}). 

\begin{table*}[t]
    \centering
    \small
    \setlength{\tabcolsep}{4.0pt}
    \begin{tabular}{llrrrrrrrrrrrrrrr}
    \toprule
     & & & \multicolumn{4}{c}{\textbf{Challenge}} & & \multicolumn{4}{c}{\textbf{Approach}} & & \multicolumn{4}{c}{\textbf{Outcome}} \\
    \cmidrule{4-7} \cmidrule{9-12} \cmidrule{14-17}
    \multicolumn{2}{l}{\textbf{Model}} & \textbf{Extractor} & \multicolumn{1}{c}{\textbf{R-1}} & \multicolumn{1}{c}{\textbf{R-2}} & \multicolumn{1}{c}{\textbf{R-L}} & \multicolumn{1}{c}{\textbf{BS}} & & \multicolumn{1}{c}{\textbf{R-1}} & \multicolumn{1}{c}{\textbf{R-2}} & \multicolumn{1}{c}{\textbf{R-L}} & \multicolumn{1}{c}{\textbf{BS}} & & \multicolumn{1}{c}{\textbf{R-1}} & \multicolumn{1}{c}{\textbf{R-2}} & \multicolumn{1}{c}{\textbf{R-L}} & \multicolumn{1}{c}{\textbf{BS}} \\
    \midrule
    \multicolumn{2}{l}{\textbf{Lead-1}} & \multicolumn{1}{l}{\textbf{-}} & 26.86 & 9.48 & 20.61 & 0.604 & & 23.23 & 7.68 & 18.31 & 0.589 & & 16.92 & 2.50 & 12.19 & 0.583 \\
    \multicolumn{2}{l}{\textbf{TextRank}} & \multicolumn{1}{l}{\textbf{-}} & 18.47 & 2.43 & 13.42 & 0.572 & & 19.29 & 4.79 & 13.87 & 0.591 & & 16.86 & 2.79 & 11.91 & 0.594 \\
    \midrule
    \multirow{5}{*}{\rotatebox[origin=c]{90}{\textbf{$\text{BART}_{\textsc{BASE}}$}}} & \multirow{4}{*}{\textbf{EtA}} & \multicolumn{1}{l}{\textbf{$\text{ST5}_{\textsc{BASE}}$}} & 18.47 & 2.43 & 13.42 & 0.617 & & 43.31 & 19.88 & 35.61 & 0.730 & & 41.39 & 19.70 & 34.86 & 0.723 \\
    & & \multicolumn{1}{l}{\textbf{$\text{ST5}_{\textsc{LARGE}}$}} & 19.02 & 2.57 & 13.86 & 0.620 & & 44.51 & 21.30 & 37.71 & 0.739 & & 39.47 & 18.40 & 33.26 & 0.719 \\
    & & \multicolumn{1}{l}{\textbf{$\text{ST5}_{\textsc{XL}}$}} & 19.00 & 2.31 & 13.76 & 0.622 & & 45.12 & 21.17 & 37.76 & 0.739 & & 39.82 & 18.83 & 34.17 & 0.722 \\
    & & \multicolumn{1}{l}{\textbf{Gold}} & 18.66 & 2.79 & 13.68 & 0.618 & & \textbf{45.70} & \textbf{22.17} & \textbf{38.52} & \textbf{0.741} & & \textbf{45.59} & \textbf{21.73} & \textbf{37.15} & \textbf{0.739} \\
    \cmidrule{2-17}
    & \textbf{E2E} & \multicolumn{1}{l}{\textbf{-}} & \textbf{21.59} & \textbf{3.88} & \textbf{15.63} & \textbf{0.627} & & 42.98 & 18.75 & 35.72 & 0.728 & & 37.89 & 16.33 & 31.69 & 0.709 \\
    \midrule
    \multirow{5}{*}{\rotatebox[origin=c]{90}{\textbf{$\text{BART}_{\textsc{LARGE}}$}}} & \multirow{4}{*}{\textbf{EtA}} & \multicolumn{1}{l}{\textbf{$\text{ST5}_{\textsc{BASE}}$}} & 18.35 & 2.15 & 13.03 & 0.611 & & 44.55 & 21.44 & 37.67 & 0.732 & & 40.30 & 19.28 & 33.97 & \textbf{0.719} \\
    & & \multicolumn{1}{l}{\textbf{$\text{ST5}_{\textsc{LARGE}}$}} & 19.61 & 2.51 & 13.95 & 0.614 & & 44.21 & 20.69 & 37.11 & 0.731 & & 38.14 & 17.70 & 31.98 & 0.708 \\
    & & \multicolumn{1}{l}{\textbf{$\text{ST5}_{\textsc{XL}}$}} & 12.77 & 1.59 & 9.43 & 0.597 & & 43.88 & 20.48 & 37.09 & 0.677 & & 39.18 & 18.45 & 33.43 & 0.665 \\
    & & \multicolumn{1}{l}{\textbf{Gold}} & 17.18 & 2.27 & 12.58 & 0.585 & & \textbf{47.84} & \textbf{23.94} & \textbf{40.14} & 0.674 & & \textbf{42.82} & \textbf{20.62} & \textbf{35.17} & 0.666 \\
    \cmidrule{2-17}
    & \textbf{E2E} & \multicolumn{1}{l}{\textbf{-}} & \textbf{20.00} & \textbf{3.76} & \textbf{14.92} & \textbf{0.623} & & 44.95 & 21.82 & 38.27 & \textbf{0.741} & & 36.22 & 15.94 & 30.61 & 0.699 \\
    \midrule
    \multirow{5}{*}{\rotatebox[origin=c]{90}{\textbf{$\text{T5}_{\textsc{BASE}}$}}} & \multirow{4}{*}{\textbf{EtA}} & \multicolumn{1}{l}{\textbf{$\text{ST5}_{\textsc{BASE}}$}} & 18.93 & 2.45 & 13.56 & 0.610 & & 44.82 & 22.40 & 38.36 & 0.732 & & 42.25 & 21.72 & 34.98 & 0.721 \\
    & & \multicolumn{1}{l}{\textbf{$\text{ST5}_{\textsc{LARGE}}$}} & 18.47 & 2.35 & 13.30 & 0.609 & & 45.10 & 22.40 & 38.76 & 0.735 & & 42.10 & 21.28 & 34.56 & 0.722 \\
    & & \multicolumn{1}{l}{\textbf{$\text{ST5}_{\textsc{XL}}$}} & 18.19 & 2.31 & 12.99 & 0.592 & & 45.97 & 23.32 & 39.80 & 0.668 & & 40.27 & 19.23 & 33.30 & 0.639 \\
    & & \multicolumn{1}{l}{\textbf{Gold}} & 19.13 & 2.62 & 13.58 & 0.583 & & \textbf{47.78} & \textbf{25.07} & \textbf{40.96} & 0.671 & & \textbf{46.60} & \textbf{24.51} & \textbf{38.49} & 0.653 \\
    \cmidrule{2-17}
    & \textbf{E2E} & \multicolumn{1}{l}{-} & \textbf{21.32} & \textbf{4.75} & \textbf{15.84} & \textbf{0.623} & & 47.39 & 24.51 & 40.79 & \textbf{0.746} & & 42.27 & 21.53 & 35.81 & \textbf{0.724} \\
    \midrule
    \multirow{5}{*}{\rotatebox[origin=c]{90}{\textbf{$\text{T5}_{\textsc{LARGE}}$}}}  & \multirow{4}{*}{\textbf{EtA}} & \multicolumn{1}{l}{\textbf{$\text{ST5}_{\textsc{BASE}}$}} & 19.18 & 2.91 & 14.04 & 0.609 & & 45.93 & 23.21 & 39.18 & 0.736 & & 42.88 & 22.30 & 35.89 & 0.726 \\
    & & \multicolumn{1}{l}{\textbf{$\text{ST5}_{\textsc{LARGE}}$}} & 18.80 & 2.96 & 13.59 & 0.612 & & 45.51 & 22.74 & 39.24 & 0.738 & & 41.76 & 21.52 & 35.71 & \textbf{0.724} \\
    & & \multicolumn{1}{l}{\textbf{$\text{ST5}_{\textsc{XL}}$}} & 19.24 & 2.76 & 13.68 & 0.591 & & 46.30 & 23.56 & 40.22 & 0.666 & & 42.32 & 21.55 & 35.94 & 0.652 \\
    & & \multicolumn{1}{l}{\textbf{Gold}} & 19.37 & 3.01 & 13.85 & 0.591 & & \textbf{49.05} & \textbf{26.04} & \textbf{42.30} & 0.669 & & \textbf{46.57} & \textbf{24.39} & \textbf{38.21} & 0.659 \\
    \cmidrule{2-17}
    & \textbf{E2E} & \multicolumn{1}{l}{\textbf{-}} & \textbf{21.17} & \textbf{4.92} & \textbf{16.13} & \textbf{0.626} & & 46.95 & 23.67 & 40.41 & \textbf{0.743} & & 41.41 & 20.90 & 34.79 & 0.721 \\
    \bottomrule
    \end{tabular}
    \caption{Performance of EtA and E2E models (best results for each aspect and metric are \textbf{bolded}).}
    \label{ta:summarization}
\end{table*}

Table \ref{ta:summarization} shows the results for both the EtA and E2E approaches. On the aspects of Approach and Outcome, EtA outperforms E2E when the gold extraction labels are used. For the $\text{BART}_{\textsc{BASE}}$ model, even when predictions from an extractive model ($\text{ST5}_{\textsc{XL}}$) are used instead of gold labels, the two-stage approach outperforms the end-to-end approach thus indicating that the two-staged approach suffers from error propagation when there is only a weak extractive model available.

\begin{figure}[t]
    \small
    \centering
    \includegraphics[width=0.8\columnwidth]{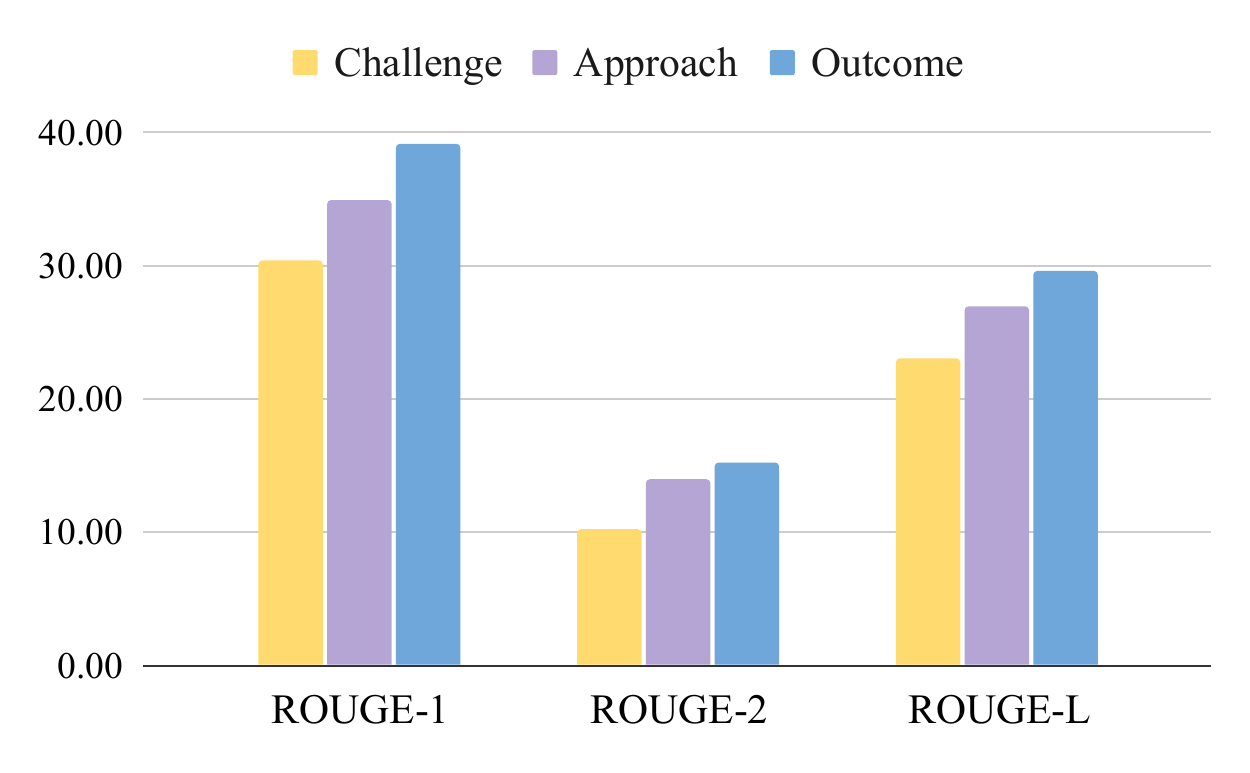}
    \caption{Average ROUGE scores between aspect-annotated sentences and abstractive summaries.}
    \label{fig:highlight_rouge}
\end{figure}
\begin{figure}[t]
    \small
    \centering
    \includegraphics[width=0.8\columnwidth]{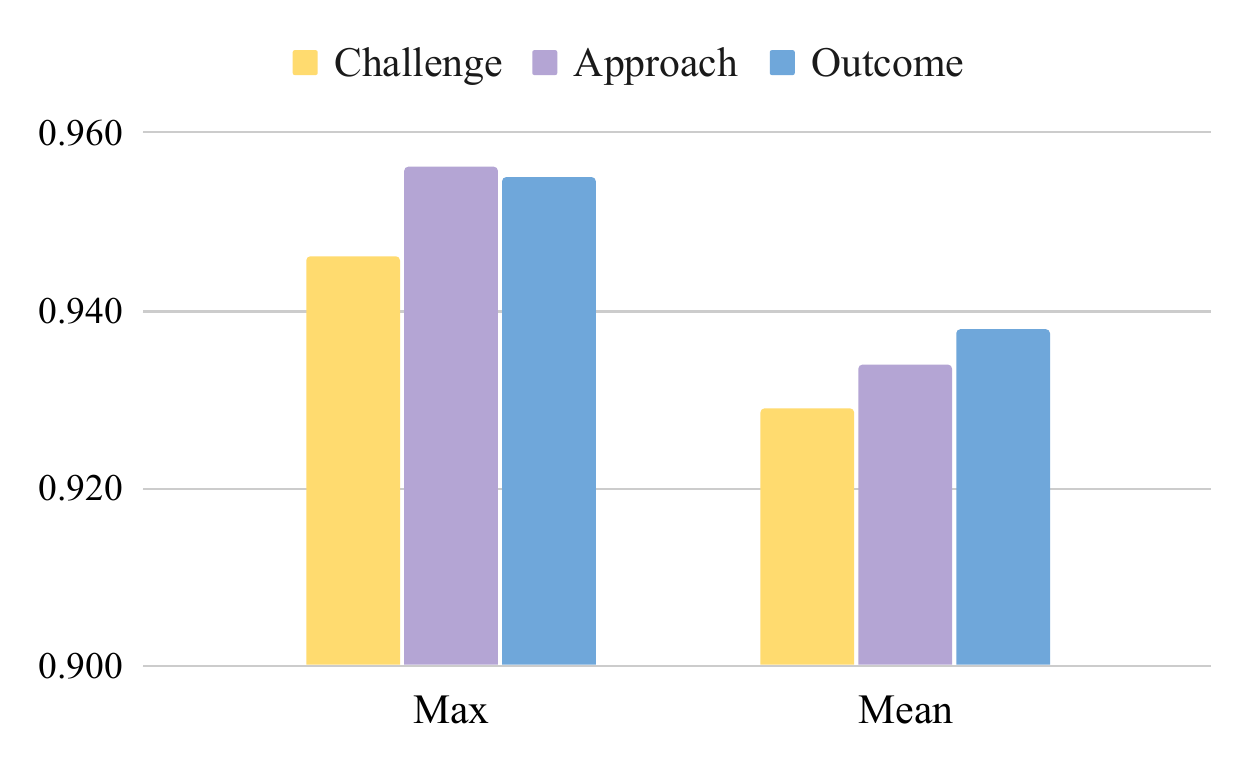}
    \caption{Maximum and average similarity of each aspect-annotated sentence to the centroid of sentences for that aspect using Sentence-T5 embeddings.}
    \label{fig:highlight_similarities}
\end{figure}

All models perform substantially worse on the Challenge aspect, even being outperformed by a simple baseline (Lead-1). To better understand this result, we question the degree to which summarization models are required to synthesize relevant sentences from a source document, that is: how much abstraction is needed? To answer this question, we perform three lines of analysis. We first compute the average ROUGE scores between each sentence annotated as a relevant aspect and the corresponding abstractive summary. Higher scores indicate that aspect-relevant sentences contain more information required to form a summary, i.e., less abstraction is needed. Results in Figure \ref{fig:highlight_rouge} show that the Challenge aspect demands models to perform a higher-level abstraction compared to the other two aspects, making it the most challenging aspect in our dataset. We next compute for each aspect the maximum and average similarity of each aspect-annotated sentence to the centroid of sentences for that aspect in the document using sentence embeddings obtained by the Sentence-T5 (-base) model. Higher numbers indicate that relevant sentences are semantically similar to each other. The results in Figure \ref{fig:highlight_similarities} show that, indeed, the relevant sentences in the Challenge portion are semantically more scattered than other aspects. This asks models to merge more semantically dissimilar sentences to produce a final summary. Lastly, we compute the entropy over the appearance of 1000 most frequent words in aspect-relevant sentences for each aspect and find that the distribution of the Challenge has the highest entropy (Challenge: 9.39, Approach: 9.21, Outcome: 9.06). This indicates that models have fewer cues for aspect-relevant sentences for the Challenge than others making it more challenging to detect relevant information from the source documents. Together, these findings indicate that the Challenge aspect of our dataset is harder to capture in summaries because models are required to perform a higher level of abstraction over sentences dissimilar to one another with fewer cues.

\begin{table}[t]
    \small
    \centering
    \begin{tabularx}{\columnwidth}{ | X | }
    \hline
    \textbf{Instruction} \\
    \hline
    Generate the indices of the sentences in the given research paper that are relevant to the paper's challenge, and then summarize them into one sentence.\\
    \hline
    \textbf{Input} \\
    \hline
    0: In this paper, we explore correlation... 1: Using the correlation measure... 2: Different from previous studies, we propose an... 3: The correlations are further [...] \\
    \hline
    \textbf{Output} \\
    \hline
    Index: 17, 18, 19, 20, 22, 23, 24 \\
    Summary: A generally accessible NER system for QA systems produces a larger answer candidate set which would be hard for current surface word-level ranking methods.\\
    \hline
    \end{tabularx}
    \caption{A training sample for EtA-CoT tuning.}
    \label{ta:eta-cot}
\end{table}

\subsection{RQ2: CoT vs. E2E instruct-tuning.}
Next, we take the popular Llama 2 model \citep{touvron_llama_2023} as a representative of recently proposed LLMs to evaluate its summarization abilities using \datasetname by fine-tuning it in two different ways. The first strategy simply fine-tunes the model to generate a summary given an instruction (E2E). In contrast, the second strategy, dubbed extract-then-abstract chain-of-thought (EtA-CoT), trains the model to generate a summary by first generating a list of indexes to sentences that are relevant to produce the summary as an immediate reasoning step \citep{wei2022chain}. We build this instruction-tuning dataset using our extractive and abstractive summarization annotations.

\paragraph{Experimental setup.}
We follow the training scheme used by \citet{alpaca}: we apply LoRA \citep{hu2022lora} and enable gradient checkpointing to fine-tune the Llama 2 7B. We train one model on a joint dataset of all three aspects and specify the target aspect in the instruction. We only trained the Llama 2 model but not its instruction-tuned variant since in our preliminary study we only observed marginal differences between them.  A training data sample used for EtA-CoT is shown in Table \ref{ta:eta-cot}. We keep the batch size to 1, number of input tokens to 4500, and test for learning rate $\in$ \{1e-4, 3e-4, 5e-4\}. We use the validation split to find the best hyperparameter and report the average performance of three differently seeded models at test time. We also report results by zero-shot prompting using the instruction-tuned Llama 2 Chat model.

\paragraph{Results and discussions.}
\begin{table}[t]
    \centering
    \small
    \setlength{\tabcolsep}{3.3pt}
    \begin{tabular}{clrrrr|r}
    \toprule
     & & \multicolumn{1}{c}{\textbf{R-1}} & \multicolumn{1}{c}{\textbf{R-2}} & \multicolumn{1}{c}{\textbf{R-L}} & \multicolumn{1}{c}{\textbf{BS}} & \multicolumn{1}{|c}{\textbf{F1}} \\
    \midrule
    \multirow{3}{*}{\textbf{Challenge}} & \textbf{Zero-Shot} & 21.37 & 5.39 & 14.55 & 0.61 & \multicolumn{1}{r}{-} \\
     & \textbf{E2E} & \textbf{30.06} & \textbf{11.33} & \textbf{23.87} & \textbf{0.67} & \multicolumn{1}{r}{-} \\
     & \textbf{EtA-CoT} & 12.48 & 4.84 & 9.25 & 0.48 & 15.9 \\
    \midrule
    \multirow{3}{*}{\textbf{Approach}} & \textbf{Zero-Shot} & 29.25 & 11.77 &21.72 & 0.66 & \multicolumn{1}{r}{-} \\
     & \textbf{E2E} & \textbf{44.01} & \textbf{23.03} & \textbf{38.58} & \textbf{0.73} & \multicolumn{1}{r}{-} \\
     & \textbf{EtA-CoT} & 26.59 & 13.15 & 22.46 & 0.59 & 10.0 \\
    \midrule
    \multirow{3}{*}{\textbf{Outcome}} & \textbf{Zero-Shot} & 27.89 & 11.12 & 20.47 & 0.65 & \multicolumn{1}{r}{-} \\
     & \textbf{E2E} & \textbf{32.85} & \textbf{13.39} & \textbf{27.23} & \textbf{0.68} & \multicolumn{1}{r}{-} \\
     & \textbf{EtA-CoT} & 23.85 & 11.03 & 20.03 & 0.57 & 5.1 \\
    \bottomrule
    \end{tabular}
    \caption{Performance of Llama 2 when trained on our dataset (E2E or EtA-CoT) and Llama 2 Chat with zero-shot prompting.}
    \label{ta:llama}
\end{table}

Table \ref{ta:llama} presents ROUGE scores, BERTScores, and extraction performance using our gold extractive labels measured by F1 for EtA-CoT models. While it is difficult to compare performance between Llama 2 and T5 due to the massive difference in model sizes (Llama 2: 7B vs. $\text{T5}_{\textsc{LARGE}}$: 770M parameters), the E2E model with Llama 2 substantially outperforms the latter on the Challenge aspect. However, it performs comparably to PLMs-based models on the other two aspects. Between the two training strategies, the E2E outperforms EtA-CoT, although it receives an additional extractive training signal during training. Poor F1 scores indicate they fail at the extraction stage, and the errors propagate to the abstraction stage. To see if models' outputs contain valid sequences for labels, we compute the average success rates by checking if (1) models predict at least one index to a sentence and (2) predicted indexes are in the valid range. We observe that 99\% of the outputs successfully fruitful both criteria. This result shows that the models have learned the required output structure yet perform poorly on prediction. By comparing zero-shot prompting and end-to-end tuning, one can observe that even with LLMs that are shown to be strong in summarization without any training, our dataset can help to improve their performance.

\subsection{RQ3: How good is the heuristic for inducing extractive summarization labels?}
\begin{table}[t]
    \centering
    \small
    \setlength{\tabcolsep}{2.5pt}
    \begin{tabular}{lrrrrrrrrrrr}
    \toprule
     & \multicolumn{3}{c}{\textbf{Challenge}} & & \multicolumn{3}{c}{\textbf{Approach}} & & \multicolumn{3}{c}{\textbf{Outcome}} \\
    \cmidrule{2-4} \cmidrule{6-8} \cmidrule{10-12}
    \textbf{Type} & \multicolumn{1}{c}{\textbf{P}} & \multicolumn{1}{c}{\textbf{R}} & \multicolumn{1}{c}{\textbf{F1}} & & \multicolumn{1}{c}{\textbf{P}} & \multicolumn{1}{c}{\textbf{R}} & \multicolumn{1}{c}{\textbf{F1}} & & \multicolumn{1}{c}{\textbf{P}} & \multicolumn{1}{c}{\textbf{R}} & \multicolumn{1}{c}{\textbf{F1}}\\
    \midrule
    \textbf{R-1}  & 80.7 & 66.1 & 69.0 & & 82.7 & 61.1 & 63.4 & & 82.6 & 63.9 & 66.9 \\
    \textbf{R-2} & \textbf{82.3} & \textbf{66.7} & \textbf{70.1} & & \textbf{84.7} & \textbf{61.2} & \textbf{63.5} & & 82.9 & \textbf{64.5}  & 67.1 \\
    \textbf{R-L} & 80.2 & 64.2 & 66.9 & & 85.7 & 60.0 & 61.8 & & \textbf{84.1} & 64.2 & \textbf{67.4}\\
    \bottomrule
    \end{tabular}
    \caption{Performance of greedy algorithm. The numbers by the best performing ROUGE function are \textbf{bolded}.}
    \label{ta:heuristic}
\end{table}

Existing works on extractive summarization systems use silver labels induced by a heuristic algorithm to work around the lack of ground-truth annotations for extractive summarization \citep{nallapati2017summarunner,narayan_dont_2018} by producing extractive labels given a source document and the corresponding abstractive summary. While this approach has been the \textit{de facto} standard \citep{liu-lapata-2019-text,pilault-etal-2020-extractive,hsu-etal-2018-unified}, no previous work assessed the quality of the heuristically induced labels against manually annotated gold labels because such evaluation would require a dataset, like \datasetname, with both extractive and abstractive annotations.

We use the greedy algorithm proposed by \citet{nallapati2017summarunner}, which induces extractive labels by adding one sentence from the source document which maximizes the ROUGE score of the set of selected sentences w.r.t. the abstractive reference summary at each iteration until no remaining sentence can improve the ROUGE scores. The resulting set of selected sentences then is used as labels for extractive summarization. We run this algorithm over our dataset and evaluate the induced extractive labels against our manually annotated gold labels. Because some of the existing works do not explicitly mention the ROUGE function used to select the sentences, we compare the three common variants.
Results are shown in Table \ref{ta:heuristic}. The best F1 score across ROUGE types and portions in our dataset is $70.1$ which arguably indicates the rather low quality of the extractions produced by this method when compared to a human gold standard.
To assess the quality of the silver labels as training data, we re-run the experiments with extract-then-abstract (EtA) pipelines in Section \ref{sec:rq1} by training extractive models on the silver labels instead of manually annotated gold labels. While pipelines with extractive models trained on gold labels outperform their counterparts trained on silver labels, the gaps are marginal. This result indicates that even though, manually created gold labels are preferred for accurate evaluations however silver labels would be sufficient for training purpose. Table \ref{ta:silver-vs-gold} in the Appendix shows the full result.
\section{Conclusion}
In this paper, we presented \datasetname, a manually crafted and validated multi-aspect summarization dataset for both extractive and abstractive summarization systems. Using \datasetname, we performed experiments using summarization models based on pretrained language models and more recent large language models such as Llama, as well as evaluating a standard algorithm to create extractive summarization datasets. In future work, we plan to explore ways to use our annotated data to bootstrap and extend our dataset through (semi-)automatic data augmentation methods, as well as build datasets for other fields, including other areas of Computer Science and other domains, possibly in languages other than English, such as, e.g., publications from the social sciences and humanities. We additionally plan to explore ways to use our aspect-based single document summarization models to enable multi-document summarization of scientific publications, a yet under-researched setup with much potential to provide challenging tasks in the age of large-scale text understanding and generation. 
\section{Limitations}
Our dataset is limited in two ways. Due to the difficulty of the annotation process, which needs to rely on experts in the scholarly domain, it contains only one reference summary for each document and aspect, and fewer samples compared to (semi)-automatically generated datasets. Moreover, we focus this initial contribution on scientific publications from a single field and language, namely English NLP papers from the ACL Anthology. 


\bibliography{refs,anthology}

\appendix

\section{Appendix}
\label{sec:appendix}
We provide our full annotation guideline for the dataset creation task in Section \ref{sec:summary_guideline} and dataset validation task in Section \ref{sec:validation_guideline}. Table \ref{tab:model-list} and \ref{tab:package-list} list models and software used in our study with external URLs, respectively.

\subsection{Annotation Guideline for Multi-aspect Summarization Dataset}
\label{sec:summary_guideline}

\subsubsection{Background}

While there are a number of datasets for the "single-document summarization" task where one source document is coupled with one generic summary, there is no dataset created from scratch for "multi-aspect summarization" where there are multiple summaries for a document focusing on different aspects.

In this annotation task, we aim to construct a dataset where there are three one-sentence summaries about \textbf{challenge}, \textbf{approach}, and \textbf{outcome} for one research article.

\subsubsection{Task Description}

The goal of this annotation task is to construct a dataset for multi-aspect summarization systems where one source document is coupled with summaries that each focus on different aspects in the source document.
We work with documents from the scholarly domain, i.e., our source documents are academic research papers. Specifically, we annotate papers that have been published in major NLP conferences (ACL, NAACL, EMNLP, EACL, AACL) and the aspects of interests are \textbf{CHALLENGE}, \textbf{APPROACH} and \textbf{OUTCOME}.

For defining each aspect, we take a subset of the categories proposed in \citet{fisas_discoursive_2015} and make small wording modifications, shown as follows:

\begin{itemize}[itemsep=0mm,leftmargin=7mm]
    \item \textbf{CHALLENGE}: The current situation faced by the researcher; it will normally include a Problem Statement, the Motivation, a Hypothesis and/or a Goal.
    \item \textbf{APPROACH}: How they intend to carry out the investigation, comments on a theoretical model or framework.
    \item \textbf{OUTCOME}: Overall conclusion that should reject or support the research hypothesis.
\end{itemize}

\subsubsection{Data}

We sample 1000 papers from the ACL anthology and use them as target documents for our annotation. All of the selected papers are from ACL, NAACL, EMNLP, EACL, and AACL.

\subsubsection{Annotation Platform}

We use INCEpTION \citep{klie-etal-2018-inception} to perform our annotation task.

\subsubsection{Annotation Procedure}

\paragraph{Step 0: Open a document}

Open a new document in INCEpTION to start the annotation.

\paragraph{Step 1: Understand the document}

Skim through the title (in the spreadsheet), abstract, introduction and conclusion of the document to identify the "main contribution" of the paper. If you find too many PDF parsing errors at this stage, skip the document by just moving on to the next one.

There may be several pages for all the lines, we suggest to increase the "Page size" to 1000 so you can see all the lines in one pages.
To do that, with one document opened, you click the "gear" button on top, and increase the number in the "Page size" form.

And we also recommend to change the color scheme for highlights to "dynamic pastelle" in the same configuration page to ease distinguishing highlights for different aspects.

\paragraph{Step 2: Read and highlight relevant text sequences}

\begin{figure}
    \centering
    \includegraphics[width=0.9\columnwidth]{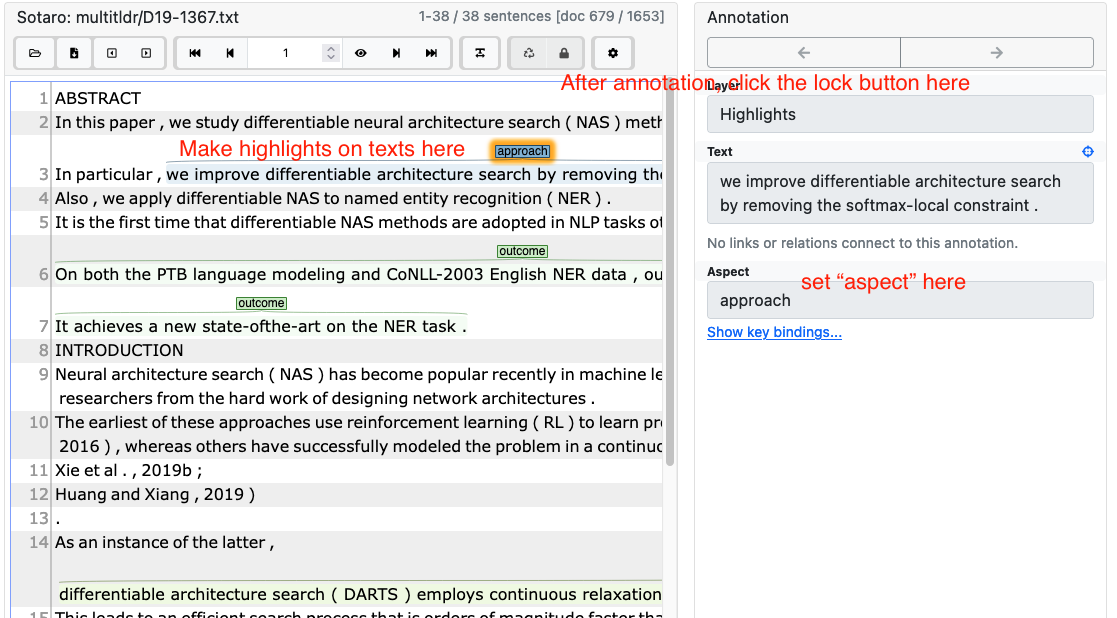}
    \caption{Screenshot of annotation procedure with INCEpTION.}
    \label{fig:inception-screenshot}
\end{figure}

Read again but sentence by sentence the abstract, introduction and conclusion, and highlight text passages (not necessarily an entire sentence) that should be included in final summary using the INCEpTION highlighting tool.
After selecting a substring for highlighting, use the `Aspect` section in the right sidebar to assign the corresponding aspect type to the highlighted text passage. Type `c` for CHALLENGE, `a` for APPROACH and `o` for OUTCOME. Highlight the same information multiple times if it is relevant and appears multiple times with/without different wording.

Some points

\begin{itemize}[itemsep=0mm,leftmargin=7mm]
    \item Make sure the highlighted sentences are relevant to the "main contribution" identified in the Step 1
    \item You will later fuse highlighted sentences into a summary and not all the information in the sentences need to be included
\end{itemize}

\paragraph{Step 3: Write summaries}

Review all the highlights aggregated in the sidebar for each aspect, and create one-sentence summary for each of them.
Pack as much as information possible in the word limitation ($<= 25$ words) for each summary.
Remove the highlighting if the information is not included in the final summary.
Save the summaries in the corresponding row in the spreadsheet.

Each summary needs to fulfill the following constraints:

\begin{itemize}[itemsep=0mm,leftmargin=7mm]
    \item Each summary contains only one sentence, and has less than 25 words
    \item Each summary cannot reference other summaries on different aspects (an example below)
    \item All the information can be found in the considered sections
\end{itemize}

\subsubsection{FAQ}

\paragraph{Is the mention of a newly created dataset APPROACH or OUTCOME?}

If authors discuss findings based on experiments using the dataset, highlight it as an OUTCOME.
If they discuss how and why the dataset has been created, highlight it as an APPROACH.

\paragraph{Is it possible for one sentence to have both aspects?}

Yes, for instance, there may be a sentence which mentions, CHALLENGE and APPROACH.
In this case, highlight the text separately for both aspects in the same sentence, possibly with the some overlaps, and make sure to provide a correct `tag` for each in the sidebar.

\paragraph{Is "Concluding remarks" same as "Conclusion"?}

Yes, we consider them to be the same.

\subsection{Annotation Guideline for Validating Summaries}
\label{sec:validation_guideline}

\subsubsection{Background}

The aim of this work is to create multi-document summarization datasets with highlight annotations.
The resulting dataset will have gold standards that can be used for development and evaluation for both abstractive and extractive summarization systems.

Since, we build our dataset using research papers from ACL conferences, the experts with the domain knowledge are required to validate the quality of this new dataset.

\subsubsection{Dataset}

Each data sample is composed of two kinds of annotations.

- \textbf{Highlight} for relevant sentences
- One-sentence \textbf{Summary} which merges the highlighted sentences

Since, this dataset is \textbf{multi}-document summarization dataset, we have both kinds of annotations for 3 aspects, namely,

\begin{itemize}[itemsep=0mm,leftmargin=7mm]
    \item \textbf{Challenge}: The current situation faced by the researcher; it will normally include a Problem Statement, the Motivation, a Hypothesis and/or a Goal.
    \item \textbf{Approach}: How they intend to carry out the investigation, comments on a theoretical model or framework.
    \item \textbf{Outcome}: Overall conclusion that should reject or support the research hypothesis.
\end{itemize}

Overall, for one research paper, there are 3 sets of highlights and 3 one-sentence summaries.

\subsubsection{Task Description}

Your task for this annotation project is to validate the quality of summaries according to the following three criteria:

\begin{itemize}[itemsep=0mm,leftmargin=7mm]
    \item \textbf{Relevance}: measures how well the summary captures the key points of the source document. If you find multiple key points in the source document, check if the most important one is included in the summary. The summaries may not contain all the key points due to the length limitation (less than 25 words per summary).
    \item \textbf{Consistency (Faithfulness)}: measures if the facts in the summary are consistent with the facts in the source document. See the highlighted sentences of the corresponding aspect in the source document and check whether the summary does reproduce all facts accurately and does not make up untrue information.
    \item \textbf{Fluency}: measures the quality of the summary as a sentence. Check if they are well-written and grammatically correct. 
\end{itemize}

\subsubsection{Annotation Procedure}

In our annotation task, we only use the following sections of a paper instead of the entire document.

\begin{itemize}[itemsep=0mm,leftmargin=7mm]
    \item Title
    \item Abstract
    \item Introduction
    \item Conclusion (if does not exist, we use Discussion)
\end{itemize}

Read only these parts of the paper when working on the annotation task described in the following steps. 

\paragraph{Step 0: Open a document}

Open the corresponding URL to the highlighted PDF file from the spreadsheet, and check if the file matches the item that you are evaluating in spreadsheet.

\paragraph{Step 1: Evaluate Relevance}

Read the document, and identify the key points regarding to the aspect of \textbf{Challenge} in the paper.
If there are multiple, consider the most important one.

Then, read the summary about the \textbf{Challenge} in the spreadsheet, and check if the key point identified in the source document is mentioned in the summary as well.

Give the scores from 1 to 5 as the following:

\begin{itemize}[itemsep=0mm,leftmargin=7mm]
    \item The summary does not include any information
    \item The summary contains some information but it is not relevant
    \item The summary contains few points that but they do not convey the main concept of the paper
    \item The summary contains key point(s) but the most important one is missing
    \item The summary contains the most important key point correctly
\end{itemize}

Then, repeat this for other two aspects, \textbf{Approach} and \textbf{Outcome}.

\paragraph{Step 2: Evaluate Consistency (Faithfulness)}

Read the summary about the \textbf{Challenge}, and check if the facts mentioned in this summary actually appears in the source document as well. In this step, you do not have to read all the source documents but only the sentences highlighted in the color which corresponds to the \textbf{Challenge} aspect.

Give the scores from 1 to 5 as the following:

\begin{itemize}[itemsep=0mm,leftmargin=7mm]
    \item The summary contains a number of critical untrue information which can critically mislead readers
    \item The summary contains few critical untrue information which can mislead readers 
    \item The summary contains some minor untrue information
    \item The summary does not contain any untrue information but readers make wrong interpretations 
    \item The summary does not contain any untrue information according to the paper and there is no space for readers to misunderstand
\end{itemize}

Then, repeat this for other two aspects, \textbf{Approach} and \textbf{Outcome}.

\paragraph{Step 3: Evaluate Fluency}

Read the summary about the \textbf{Challenge}, and check if it is well-written and grammatically correct.
In this step, you do not have to read the source document at all.

Give the scores from 1 to 5 as the following:

\begin{itemize}[itemsep=0mm,leftmargin=7mm]
    \item The summary contains a number of grammatical errors which make it unreadable
    \item The summary contains a few critical grammatical errors which lead to misunderstandings
    \item The summary contains a few minor grammatical errors which can lead to misunderstandings
    \item The summary does contain errors but they would not lead to any misunderstandings
    \item The summary does not contain any errors and there is no space for readers to misunderstand
\end{itemize}

Then, repeat this for the other two aspects, \textbf{Approach} and \textbf{Outcome}.

\newpage
\begin{table*}[t]
    \small
    \centering
    \setlength{\tabcolsep}{8.0pt}
    \begin{tabular}{lrll}
        \toprule
        \multicolumn{1}{c}{\textbf{Model}} & \multicolumn{1}{c}{\# Params} & \multicolumn{1}{c}{\textbf{Licence}} & \multicolumn{1}{c}{\textbf{URL}} \\
        \midrule
        \textbf{$\text{BART}_{\textsc{BASE}}$} & 139M & Apache 2.0 & \href{https://huggingface.co/facebook/bart-base}{https://huggingface.co/facebook/bart-base} \\
        \textbf{$\text{BART}_{\textsc{LARGE}}$} & 406M & Apache 2.0 & \href{https://huggingface.co/facebook/bart-large}{https://huggingface.co/facebook/bart-large} \\
        \textbf{$\text{T5}_{\textsc{BASE}}$} & 223M & Apache 2.0 & \href{https://huggingface.co/t5-base}{https://huggingface.co/t5-base} \\
        \textbf{$\text{T5}_{\textsc{LARGE}}$} & 738M & Apache 2.0 & \href{https://huggingface.co/t5-large}{https://huggingface.co/t5-large} \\
        \textbf{$\text{ST5}_{\textsc{BASE}}$} & 110M & Apache 2.0 & \href{https://huggingface.co/sentence-transformers/sentence-t5-base}{https://huggingface.co/sentence-transformers/sentence-t5-base} \\
        \textbf{$\text{ST5}_{\textsc{LARGE}}$} & 335M & Apache 2.0 & \href{https://huggingface.co/sentence-transformers/sentence-t5-large}{https://huggingface.co/sentence-transformers/sentence-t5-large} \\
        \textbf{$\text{ST5}_{\textsc{XL}}$} & 1.24B & Apache 2.0 & \href{https://huggingface.co/sentence-transformers/sentence-t5-xl}{https://huggingface.co/sentence-transformers/sentence-t5-xl} \\
        \textbf{$\text{Llama 2}_{\textsc{7B}}$} & 7B & LLAMA 2 License & \href{https://huggingface.co/meta-llama/Llama-2-7b-hf}{https://huggingface.co/meta-llama/Llama-2-7b-hf} \\
        \textbf{$\text{Llama 2 Chat}_{\textsc{7B}}$} & 7B & LLAMA 2 License & \href{https://huggingface.co/meta-llama/Llama-2-7b-chat-hf}{https://huggingface.co/meta-llama/Llama-2-7b-chat-hf} \\
        \bottomrule
    \end{tabular}
    \caption{A list of models with external URLs used in our study.}
    \label{tab:model-list}
\end{table*}

\begin{table*}[t]
    \footnotesize
    \centering
    \setlength{\tabcolsep}{6.0pt}
    \begin{tabular}{lll}
        \toprule
        \multicolumn{1}{c}{\textbf{Package}} & \multicolumn{1}{c}{\textbf{Licence}} & \multicolumn{1}{c}{\textbf{URL}} \\
        \midrule
        Grobid \citep{grobid} & Apache 2.0 & \href{https://github.com/kermitt2/grobid}{https://github.com/kermitt2/grobid} \\
        INCEpTION \citep{klie-etal-2018-inception} & Apache 2.0 & \href{https://github.com/inception-project/inception}{https://github.com/inception-project/inception} \\
        PyTorch \citep{paszke2019pytorch} & BSD-style & \href{https://github.com/pytorch/pytorch}{https://github.com/pytorch/pytorch} \\
        Transformers \citep{wolf-etal-2020-transformers} & Apache 2.0 & \href{https://github.com/huggingface/transformers}{https://github.com/huggingface/transformers} \\
        Lightning & Apache 2.0 & \href{https://github.com/Lightning-AI/pytorch-lightning}{https://github.com/Lightning-AI/pytorch-lightning}\\
        scikit-learn \citep{scikit-learn} & BSD 3-Clause & \href{https://github.com/scikit-learn/scikit-learn}{https://github.com/scikit-learn/scikit-learn} \\
        Spacy \citep{Honnibal_spaCy_Industrial-strength_Natural_2020} & MIT &  \href{https://github.com/explosion/spaCy/}{https://github.com/explosion/spaCy/} \\
        SentenceTransformers \citep{reimers-gurevych-2019-sentence} & Apache 2.0 & \href{https://github.com/UKPLab/sentence-transformers}{https://github.com/UKPLab/sentence-transformers} \\
        SacreRouge \citep{deutsch-roth-2020-sacrerouge} & Apache 2.0 & \href{https://github.com/danieldeutsch/sacrerouge}{https://github.com/danieldeutsch/sacrerouge} \\
        \bottomrule
    \end{tabular}
    \caption{A list of software and libraries with external URLs used in our study.}
    \label{tab:package-list}
\end{table*}

\begin{table*}[t]
    \centering
    \small
    \setlength{\tabcolsep}{2.0pt}
    \begin{tabular}{llrrrrrrrrrrrrrrr} 
    \toprule
     & & & \multicolumn{4}{c}{\textbf{Challenge}} & & \multicolumn{4}{c}{\textbf{Approach}} & & \multicolumn{4}{c}{\textbf{Outcome}} \\ 
    \cmidrule{4-7} \cmidrule{9-12} \cmidrule{14-17}
    \multicolumn{2}{c}{\textbf{Model}} & \textbf{Label type} & \multicolumn{1}{c}{\textbf{R-1}} & \multicolumn{1}{c}{\textbf{R-2}} & \multicolumn{1}{c}{\textbf{R-L}} & \multicolumn{1}{c}{\textbf{BS}} & & \multicolumn{1}{c}{\textbf{R-1}} & \multicolumn{1}{c}{\textbf{R-2}} & \multicolumn{1}{c}{\textbf{R-L}} & \multicolumn{1}{c}{\textbf{BS}} & & \multicolumn{1}{c}{\textbf{R-1}} & \multicolumn{1}{c}{\textbf{R-2}} & \multicolumn{1}{c}{\textbf{R-L}} & \multicolumn{1}{c}{\textbf{BS}} \\
    \midrule
    \multirow{6}{*}{\rotatebox[origin=c]{90}{\textbf{$\text{BART}_{\textsc{BASE}}$}}} & \multicolumn{1}{l}{\multirow{2}{*}{\textbf{$\text{ST5}_{\textsc{BASE}}$}}} & \multicolumn{1}{l}{\textbf{Gold}} & 18.47 & 2.43 & 13.42 & 0.617 & & 43.31 & 19.88 & 35.61 & 0.730 & & 41.39 & 19.70 & 34.86 & 0.723 \\
     & & \multicolumn{1}{l}{\textbf{Heuristic}} & 17.90 & 2.44 & 13.25 & 0.614 & & 44.55 & 20.60 & 37.23 & 0.735 & & 40.43 & 19.12 & 33.53 & 0.723 \\
     & \multicolumn{1}{l}{\multirow{2}{*}{\textbf{$\text{ST5}_{\textsc{LARGE}}$}}} & \multicolumn{1}{l}{\textbf{Gold}} & 19.02 & 2.57 & 13.86 & 0.620 & & 44.51 & 21.30 & 37.71 & 0.739 & & 39.47 & 18.40 & 33.26 & 0.719 \\
     & & \multicolumn{1}{l}{\textbf{Heuristic}} & 18.31 & 2.57 & 13.41 & 0.615 & & 43.35 & 20.39 & 36.98 & 0.734 & & 39.98 & 18.33 & 32.77 & 0.716 \\
     & \multicolumn{1}{l}{\multirow{2}{*}{\textbf{$\text{ST5}_{\textsc{XL}}$}}} & \multicolumn{1}{l}{\textbf{Gold}} & 19.00 & 2.31 & 13.76 & 0.622 & & 45.12 & 21.17 & 37.76 & 0.739 & & 39.82 & 18.83 & 34.17 & 0.722 \\
     & & \multicolumn{1}{l}{\textbf{Heuristic}} & 18.71 & 2.51 & 13.95 & 0.615 & & 44.90 & 20.79 & 37.91 & 0.737 & & 39.03 & 18.45 & 32.82 & 0.711 \\
    \midrule
    \multirow{6}{*}{\rotatebox[origin=c]{90}{\textbf{$\text{BART}_{\textsc{LARGE}}$}}} & \multicolumn{1}{l}{\multirow{2}{*}{\textbf{$\text{ST5}_{\textsc{BASE}}$}}} & \multicolumn{1}{l}{\textbf{Gold}} & 18.35 & 2.15 & 13.03 & 0.611 & & 44.55 & 21.44 & 37.67 & 0.732 & & 40.30 & 19.28 & 33.97 & 0.719 \\
     & & \multicolumn{1}{l}{\textbf{Heuristic}} & 19.02 & 2.36 & 13.53 & 0.607 & & 44.59 & 20.86 & 37.97 & 0.735 & & 39.48 & 18.38 & 33.68 & 0.716 \\
     & \multicolumn{1}{l}{\multirow{2}{*}{\textbf{$\text{ST5}_{\textsc{LARGE}}$}}} & \multicolumn{1}{l}{\textbf{Gold}} & 19.61 & 2.51 & 13.95 & 0.614 & & 44.21 & 20.69 & 37.11 & 0.731 & & 38.14 & 17.70 & 31.98 & 0.708 \\
     & & \multicolumn{1}{l}{\textbf{Heuristic}} & 18.64 & 2.39 & 13.50 & 0.610 & & 44.20 & 21.52 & 38.05 & 0.737 & & 40.18 & 19.47 & 34.10 & 0.720 \\
     & \multicolumn{1}{l}{\multirow{2}{*}{\textbf{$\text{ST5}_{\textsc{XL}}$}}} & \multicolumn{1}{l}{\textbf{Gold}} & 12.77 & 1.59 & 9.43 & 0.597 & & 43.88 & 20.48 & 37.09 & 0.677 & & 39.18 & 18.45 & 33.43 & 0.665 \\
     & & \multicolumn{1}{l}{\textbf{Heuristic}} & 19.12 & 2.33 & 13.31 & 0.607 & & 43.93 & 20.93 & 37.14 & 0.734 & & 38.20 & 17.32 & 32.19 & 0.712 \\
    \midrule
    \multirow{6}{*}{\rotatebox[origin=c]{90}{\textbf{$\text{T5}_{\textsc{BASE}}$}}} & \multicolumn{1}{l}{\multirow{2}{*}{\textbf{$\text{ST5}_{\textsc{BASE}}$}}} & \multicolumn{1}{l}{\textbf{Gold}} & 18.93 & 2.45 & 13.56 & 0.610 & & 44.82 & 22.40 & 38.36 & 0.732 & & 42.25 & 21.72 & 34.98 & 0.721 \\
     & & \multicolumn{1}{l}{\textbf{Heuristic}} & 19.48 & 2.86 & 14.18 & 0.607 & & 45.64 & 23.76 & 39.73 & 0.740 & & 41.19 & 21.03 & 34.40 & 0.716 \\
     & \multicolumn{1}{l}{\multirow{2}{*}{\textbf{$\text{ST5}_{\textsc{LARGE}}$}}} & \multicolumn{1}{l}{\textbf{Gold}} & 18.47 & 2.35 & 13.30 & 0.609 & & 45.10 & 22.40 & 38.76 & 0.735 & & 42.10 & 21.28 & 34.56 & 0.722 \\
     & & \multicolumn{1}{l}{\textbf{Heuristic}} & 18.56 & 2.45 & 13.56 & 0.606 & & 45.47 & 23.06 & 39.27 & 0.739 & & 40.36 & 20.31 & 33.66 & 0.707 \\
     & \multicolumn{1}{l}{\multirow{2}{*}{\textbf{$\text{ST5}_{\textsc{XL}}$}}} & \multicolumn{1}{l}{\textbf{Gold}} & 18.19 & 2.31 & 12.99 & 0.592 & & 45.97 & 23.32 & 39.80 & 0.668 & & 40.27 & 19.23 & 33.30 & 0.639 \\
     & & \multicolumn{1}{l}{\textbf{Heuristic}} & 18.31 & 2.28 & 13.08 & 0.600 & & 44.59 & 22.79 & 38.62 & 0.736 & & 40.36 & 20.31 & 33.66 & 0.707 \\
    \midrule
    \multirow{6}{*}{\rotatebox[origin=c]{90}{\textbf{$\text{T5}_{\textsc{LARGE}}$}}} & \multicolumn{1}{l}{\multirow{2}{*}{\textbf{$\text{ST5}_{\textsc{BASE}}$}}} & \multicolumn{1}{l}{\textbf{Gold}} & 19.18 & 2.91 & 14.04 & 0.609 & & 45.93 & 23.21 & 39.18 & 0.736 & & 42.88 & 22.30 & 35.89 & 0.726 \\
     & & \multicolumn{1}{l}{\textbf{Heuristic}} & 19.35 & 3.08 & 14.19 & 0.611 & & 46.33 & 23.80 & 39.98 & 0.742 & & 42.63 & 22.36 & 35.95 & 0.725 \\
     & \multicolumn{1}{l}{\multirow{2}{*}{\textbf{$\text{ST5}_{\textsc{LARGE}}$}}} & \multicolumn{1}{l}{\textbf{Gold}} & 18.80 & 2.96 & 13.59 & 0.612 & & 45.51 & 22.74 & 39.24 & 0.738 & & 41.76 & 21.52 & 35.71 & 0.724 \\
     & & \multicolumn{1}{l}{\textbf{Heuristic}} & 19.06 & 3.06 & 13.83 & 0.611 & & 44.83 & 22.10 & 38.75 & 0.734 & & 41.50 & 20.90 & 35.20 & 0.721 \\
     & \multicolumn{1}{l}{\multirow{2}{*}{\textbf{$\text{ST5}_{\textsc{XL}}$}}} & \multicolumn{1}{l}{\textbf{Gold}} & 19.24 & 2.76 & 13.68 & 0.591 & & 46.30 & 23.56 & 40.22 & 0.666 & & 42.32 & 21.55 & 35.94 & 0.652 \\
     & & \multicolumn{1}{l}{\textbf{Heuristic}} & 19.04 & 2.86 & 13.76 & 0.608 & & 45.79 & 22.79 & 39.40 & 0.739 & & 41.50 & 21.26 & 35.44 & 0.716 \\
    \bottomrule
    \end{tabular}
    \caption{Results of extractive models trained on silver and gold data in the extract-then-abstract approach.}
    \label{ta:silver-vs-gold}
\end{table*}


\end{document}